\documentclass[10pt,twocolumn,letterpaper]{article}

\usepackage{wacv}              
\usepackage{graphicx}
\usepackage{subcaption}
\usepackage[accsupp]{axessibility}

\newcommand{\new}[1]{{\color{brown}#1}}

\newcommand{\method}{AutoSew\xspace}
\newcommand{\dataset}{\textsc{GarmentCodeData.v2}\xspace}
\newcommand{\newdataset}{\textsc{M-E.GarmentCodeData}\xspace}

\definecolor{wacvblue}{rgb}{0.21,0.49,0.74}
\usepackage[pagebackref,breaklinks,colorlinks,allcolors=wacvblue]{hyperref}

\def\wacvPaperID{1283} 
\def\confName{WACV}
\def\confYear{2026}

\title{\method: A Geometric Approach to Stitching Prediction with Graph Neural Networks}

\author{
    Pablo Ríos-Navarro$^{1}$ \quad\quad
    Elena Garces$^{2}$ \quad\quad
    Jorge Lopez-Moreno$^{1}$ \\[0.2cm]
    {\tt\small pablo.rios@urjc.es  } \quad
    {\tt\small elenag@adobe.com} \quad
    {\tt\small jorge.lopez@urjc.es} \\[0.3cm]
    $^{1}$Universidad Rey Juan Carlos, Madrid, Spain \\
    $^{2}$Adobe Research, France
}

\begin{document}

\twocolumn[{%
\renewcommand\twocolumn[1][]{#1}%
\maketitle
  \centering
   \includegraphics[width=1\textwidth]{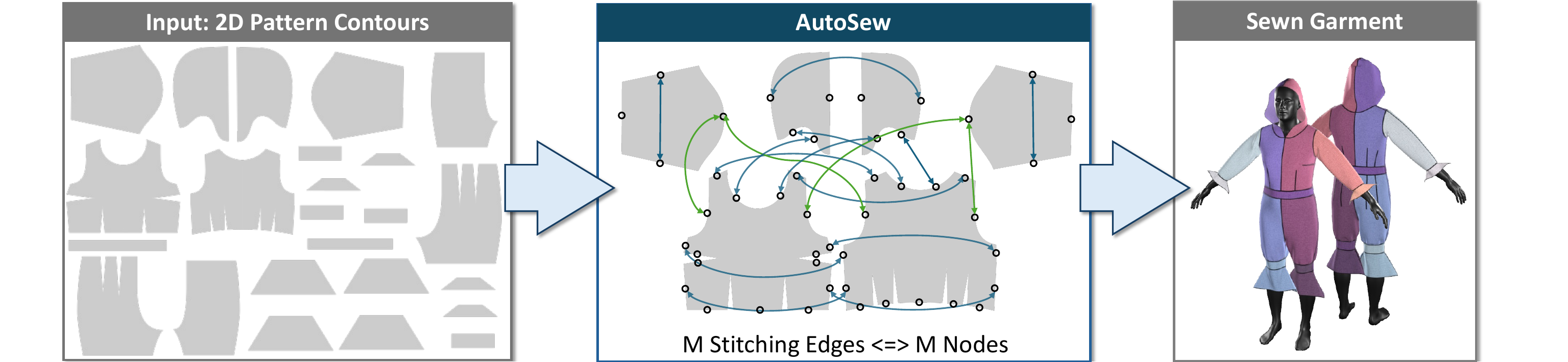}
\captionof{figure}{Left: \method takes as input a 2D sewing pattern, consisting of the contours of the panels (24 in this example) and outputs the edges that should be stitched together to assemble the garment. Middle: Our method models the stitching edges as a graph neural network and leverages feature aggregation with an adaptive matching mechanism to efficiently solve a partial assignment problem. Multi-edge stitching connections are shown in green, while one-to-one stitches are in blue. Right: 3D garment assembled from the resulting stitching connections predicted by our model.
\vspace{1.7mm}}
\label{fig:teaser}
}]

\begin{abstract}

Automating garment assembly from sewing patterns remains a significant challenge due to the lack of standardized annotation protocols and the frequent absence of semantic cues. 
Existing methods often rely on panel labels or handcrafted heuristics, which limit their applicability to real-world, non-conforming patterns. 
We present \method, a fully automatic, geometry-based approach for predicting stitch correspondences directly from 2D pattern contours. 
\method formulates the problem as a graph matching task, leveraging a Graph Neural Network to capture local and global geometric context, and employing a differentiable optimal transport solver to infer stitching relationships—including multi-edge connections. To support this task, we update the GarmentCodeData dataset modifying over 18k patterns with realistic multi-edge annotations, reflecting industrial assembly scenarios. \method achieves 96\% F1-score and successfully assembles 73.3\% of test garments without error, outperforming existing methods while relying solely on geometric input. Our results demonstrate that geometry alone can robustly guide stitching prediction, enabling scalable garment assembly without manual input. Our dataset and code are available online. \footnote{\href{https://mslab.es/projects/autosew/}{https://mslab.es/projects/autosew/}}

\end{abstract}
\section{Introduction}\label{sec:intro}


The fashion industry faces a dual challenge in its digital transformation: modernizing garment design workflows while also integrating its vast legacy of existing patterns. Traditional garment manufacturing relies on skilled tailors who create the sewing patterns and subsequently annotate them with assembly instructions to guide the manufacturing process. However, this manual annotation process heavily relies on individual experience and intuition, which often results in incomplete and non-standard annotations; hence, posing a significant obstacle to automation. 
Although pattern vectorization is widely supported by commercial tools, these systems lack guidance for stitching prediction, which remains a bottleneck for automation.
Bridging this gap—digitizing existing patterns for 3D assembly and seamless integration into digital workflows—is crucial for realizing the full potential of digital garment creation.
%
%
%
%
In this paper, we address the problem of automated garment assembly of sewing patterns, where the patterns have been digitized and vectorized but lack any semantic cue to guide the stitching process nor follow any standard.

%

%
In contrast with previous approaches, we propose \method, a novel method that leverages Graph Neural Networks (GNNs) to encode geometric features and predict stitch correspondences without relying on semantic labels or 3D information. By learning global embeddings, \method captures the essential properties of each panel edge, discerning subtle variations in edge geometry, and accurately establishing stitch connections.
Our key insight is that GNNs are ideally suited for modeling the complex relationships inherent in sewing pattern data. Unlike traditional approaches that focus on local features, GNNs effectively propagate information across the graph structure, capturing both local and global geometric context. This allows \method to robustly predict stitch correspondences in non-standardized patterns, even when local cues are ambiguous. Furthermore, the inherent invariance of GNNs to graph size and node ordering ensures seamless adaptation to diverse pattern sizes and structures.

A particularly challenging aspect of garment assembly is the presence of multi-edge stitch connections, where a single edge must be stitched simultaneously to multiple others. These configurations are common in industrial patterns—for example, sleeves are often attached to both the front and back torso panels in a one-to-many relationship. However, most existing datasets \cite{korosteleva_garmentcodedata_2024, Korosteleva_GarmentData_2021, wang20244ddress4ddatasetrealworld} and methods \cite{he_dresscode_2024, Korosteleva_NeuralTailor_2022, liu_towards_2023, Chen_panelformer_2024, nakayama_aipparel_2024} assume one-to-one stitching, which oversimplifies the actual construction logic. To tackle this problem, we have created a novel dataset, \newdataset, which includes this kind of connections.

In summary, we present the following contributions:

\begin{itemize}
	\item \newdataset, a dataset derived from \textsc{GarmentCodeData} with multi-edge stitching connections that better reflect industry-level assembly scenarios.
	
    \item A geometry-based method that automatically predicts multiple stitch correspondences in garment patterns without relying on semantic annotations. 
        
    \item A novel formulation of the problem that effectively combines a Graph Neural Network with differentiable optimal transport able to handle variable pattern sizes.     
    
    
    \item Comprehensive validation and ablation studies evaluating the contribution of each component of the proposed architecture.

\end{itemize} 

\section{Related Work}\label{sec:related}

Despite the importance of user interaction in fashion CAD \cite{Umetani2011}, very little attention has been paid to automatic sewing for 3D garment creation. Several applications in digital fashion leverage knowledge about connected sewing edges. For instance, the simulation of 3D garments is quite dependent on correct intersection-free initialization of the 2D panels in 3D space. The automatic initialization system proposed by Liu \etal.~\cite{huamin_automatic_2024} requires full stitching information as input to initialize 68\% of their garments correctly without additional user intervention. 



An earlier approach to automatic stitching was proposed by Berthouzoz \etal.~\cite{berthouzoz_parsing_2013}. Their method combines machine learning with integer programming to parse a dataset of standardized sewing patterns and stitch them into 3D garments. Although this approach can handle decorative panels and multi-edge stitching, it is largely dependant on semantic annotations such as panel names and manually placed notches —small alignment markers added to the pattern edges to guide the stitching process. In their 50-garments dataset, they reported a $68\%$ of correctly sewn full garments (GSP), and a 87\% F1-score in automatic stitching. Regrettably, they state that their approach cannot solve the stitching problem without panel-level information, and their dataset and code are not publicly available for comparison purposes.

%

One of the challenges that faces a learning-based solution is the availability of data. 
Korosteleva and Lee \cite{Korosteleva_GarmentData_2021} proposed a large dataset (20k garments) and a pattern generator pipeline capable of creating novel variations. They further improved the parametric pattern model and its generative capabilities with a hierarchical, component-oriented programming framework \cite{Korosteleva_GarmentCode_2023}, which was later used to increase their original dataset to 115k garments \cite{korosteleva_garmentcodedata_2024}, fitted to thousands of different bodies obtained from the parametric data-driven SMPL human model \cite{Loper2015}. 
%
These datasets are mostly oriented to generating synthetic dressed humans, thus, lack real-world features such as multi-edge annotations. 

The dataset from \cite{Korosteleva_GarmentData_2021} has been used in several approaches that aim to reconstruct the 3D garment from different inputs: $NeuralTailor$, created by Korosteleva et al.~\cite{Korosteleva_NeuralTailor_2022} focuses on reconstructing 2D garment sewing patterns from 3D point clouds of garment models. They use panel and stitching encoding and decoding with a combination of RNN (LSTM) and MLP modules. Liu et al. \cite{liu_towards_2023} introduced $SewFormer$, a method for recovering garment sewing patterns from single photos of dressed avatars. More recently Chen et al.~\cite{Chen_panelformer_2024} presented $Panelformer$, for reconstructing garment sewing patterns from 2D garment images.
Both $Panelformer$ and $NeuralTailor$ share a very similar stitching prediction submodule, based on a similarity matrix and a feature-space distance metric. Chen et al.~\cite{Chen_panelformer_2024} report an F1-score of 79.2\% in automatic stitching, and 65.5\% for $NeuralTailor$~\cite{Korosteleva_NeuralTailor_2022} on the same unseen test set. In fact, Nakayama \etal \cite{nakayama_aipparel_2024} demonstrated that SewFormer stitching accuracy collapses to 2.8\% when re-evaluated on the updated GarmentCodeData release \cite{korosteleva_garmentcodedata_2024}. 

Concurrent work, GarmageNet~\cite{li2025garmagenetmultimodalgenerativeframework} proposes a broader generative framework for digital garment modeling. Within this framework, they introduce GarmageJigsaw, which predicts sewing connections using both 2D and 3D information through an ensemble of neural networks, partly inspired by Jigsaw~\cite{lu2023jigsawlearningassemblemultiple}, a method for assembling fractured objects by combining edge matching and global reasoning.

These neural approaches report precision and recall only on those outputs where the number and semantic identification of both  panels and edges were predicted correctly. This biased protocol may lead to an overestimation of their stitching scores and warrants careful consideration. Furthermore, stitching is treated as just one stage within a larger reconstruction pipeline that requires previous correct classification of panels and ad-hoc labels. Industry-grade patterns do not include 3D simulated data and lack most of the semantic information (pattern labels, annotated connections or unique notches) that is required by previous work~\cite{berthouzoz_parsing_2013, li2025garmagenetmultimodalgenerativeframework}. The need to minimize such assumptions to achieve unbiased automatic stitching in real production environments motivates our purely geometric approach.

Another key consideration is that transformer-based models, despite their flexibility, rely on positional encoding to preserve the order of elements.
This supports global self‑attention, but it also makes them sensitive to changes in input size and panel ordering \cite{dosovitskiy2021}. Therefore, extending them to new pattern dimensions or topologies requires retraining those positional embeddings. 
In contrast, GNNs propagate messages along the actual graph edges, needing no sequence encoding and handling any vertex order or graph size \cite{hamilton_inductive_2018}. 
This structural inductive bias makes GNNs especially well‑suited to stitching prediction, where relationships between panels are just as important as the entities themselves.
GNNs have been applied to the problem of graph matching, that is, finding a correspondence between the nodes of two or more graphs, based on their structure or attributes~\cite{Zanfir_2018,li19,Wang_2019}. Feature matching is an interesting extension: Sarlin \etal.~\cite{Sarlin2020} use a GNN ($Superglue$) to learn to match image features by considering their spatial relationships and contextual information instead of distance metrics, formulating the problem as an optimal transport solution. More recent methods also rely on message passing, but learn directly from the output of combinatorial solvers \cite{Rolinek2020}, or Quadratic Assignment Problem (QAP) formulations \cite{{Wang2021QAP}}.

We are particularly interested in the problem of partial graph matching, a scenario where only a subset of nodes across multiple graphs have correspondences (i.e., only a few edge pairs are valid matches in a garment). Wang \etal. extend GNNs to the challenge of selecting the most relevant correspondences with a differentiable top-k discrete operator~\cite{Wang2023}, while Ratnayaka \etal. \cite{ratnayaka2025} rely on a continuous optimization approach based on optimal partial transport.

\section{\method Architecture}
\label{sec:architecture}
\begin{figure*}[ht]
  \centering
   \includegraphics[width=\textwidth]{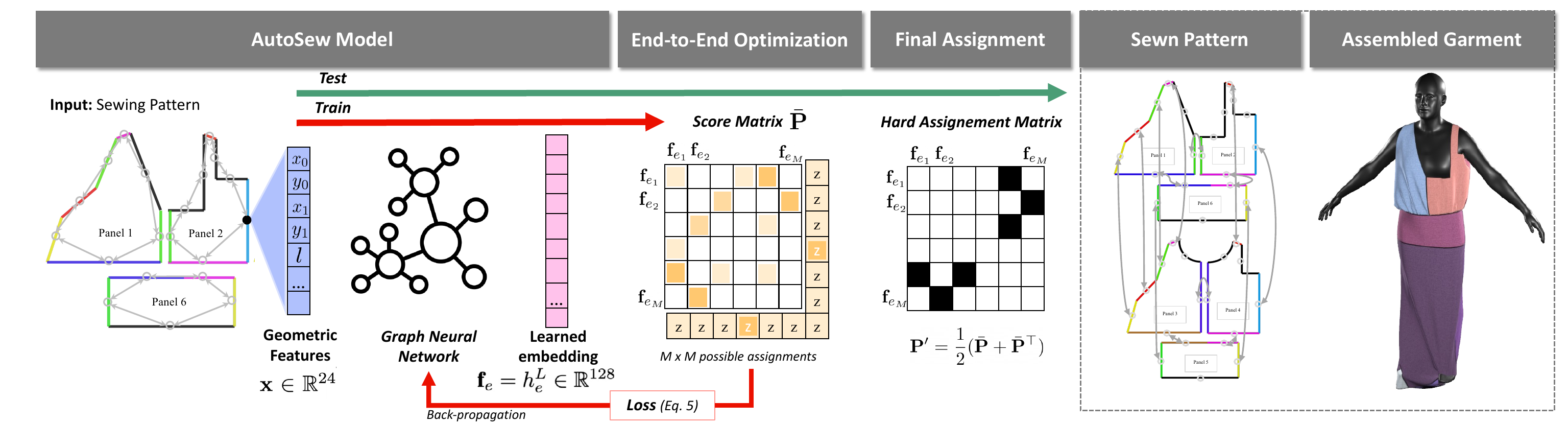}
   \caption{Overview of \method. For each sewing pattern, a graph is constructed with nodes representing stitching edges and edges capturing geometric relationships of the panels. A GNN with message passing learns embeddings for each edge, optimized for stitch correspondence using a differentiable optimal transport loss.}
   \label{fig:overview}
\end{figure*}

The complexity and variability of sewing patterns—with their diverse shapes, sizes, and structures—demand an advanced approach that accounts for both local and global information. 
To address this challenge, we formulate \method as an end-to-end graph matching problem using a novel learning-based architecture. 
For automatic pattern stitching, the task is to identify valid edge pairings while adhering to two constraints: $(1)$ each edge can be stitched to one or more edges, and $(2)$ some edges may remain unstitched.

An overview of the method is shown in Figure~\ref{fig:overview}. First, we characterize each stitching edge as a set of geometric edges features, described in Section~\ref{sec:geometry}. Then, we build a GNN where each node in the graph corresponds to a stitching edge and neighbor nodes are linked according to the 2D topology of the pattern. The graph embeddings are learned to account for non-local relationships using iterative neighbor sampling (Section~\ref{sec:graph}). Finally, we optimize the assignment between stitching edges using global optimal transport (Section~\ref{sec:optimal}). The process is learned end-to-end (Section~\ref{sec:endtoend}).

\subsection{Geometric Edge Features} \label{sec:geometry}



To characterize the edges, we extract a set of geometric features (local shape descriptors and topological properties). While state-of-the-art approaches ~\cite{berthouzoz_parsing_2013} leverage semantic information such as the panel name (front, back, sleeve,\textit{ etc.}) or edge-to-edge annotations (\textit{i.e.} notches), our method relies solely on geometric data. This design choice ensures that our model remains independent from additional semantic dependencies, thereby improving its ability to generalize across diverse sewing patterns.
We define two distinct geometric feature types:
\begin{itemize} [nosep, noitemsep]
    \item \textbf{Local shape descriptors (18 dimensions)}. These features capture the local shape of each edge, including its start and end vertices, length, orientation, curvature type, and control point parameters.

    \item \textbf{Topological properties (4 dimensions)}. These describe the contextual role of the edge within the panel, including the interior angles with its adjacent edges (left and right), the total number of edges in the panel, and a unique panel identifier.

\end{itemize}
The full list of 22 geometric features per edge is detailed in the Supplementary Material.
We then apply a preprocessing pipeline to standardize feature values: we scale all geometric values by 1/100, normalize the per-panel edge count \(N_e\) to \([0,1]\) using Min-Max scaling, and encode interior angles on both sides of each edge—left $(\sin\alpha_l,\cos\alpha_l)$ and right $(\sin\alpha_r,\cos\alpha_r)$—to ensure rotational invariance and stabilize the training \cite{zhou2020continuityrotationrepresentationsneural}. Additionally, we sort edges in anticlockwise order and translate each panel so its lower-left bounding-box corner lies at the origin. 
This yields a 24-dimensional feature vector \(\mathbf{x}\in\mathbb{R}^{24}\), which serves as input to the GNN. 

\subsection{Graph Neural Network} \label{sec:graph}

We model the 2D patterns of each garment as a Graph Neural Network \cite{scarselli_gnn_2009} where each node in the graph corresponds to a stitching edge $e$, defined by a feature vector $\textbf{x}_e$, and the connections between the nodes are created based on the topological shape of the 2D patterns. See the middle graph example in Figure \ref{fig:teaser}. 
Each graph of a garment is defined by $M$ nodes which correspond to all the available stitching edges from all panels. 
This graph representation facilitates learning long-range dependencies between stitching edges through node embeddings.  


We formulate the problem as inductive learning and employ an efficient strategy to aggregate information from non-neighbor nodes. In particular, we use GraphSAGE~\cite{hamilton_inductive_2018} which learns node embedding by a sampling and aggregation strategy. This approach allows the model to adapt to different graph structures and sizes, improving its ability to predict stitching connections in new sewing patterns.
In the GraphSAGE learning process, each node is initialized with the feature vector $\textbf{x}_e$ which is iteratively updated by aggregating features from its neighbors in $L$ hidden layers (or steps). Formally, for a node \( e \) at layer \( l \), the aggregation update can be formulated as:
%
\begin{small}
\begin{multline}
h_e^{(l+1)} = \sigma \Big( W^{(l)} \cdot \text{CONCAT} \Big( h_e^{(l)}, \\
\text{AGG} \Big( \{ h_e^{(l)} \mid e \in \mathcal{N}(e) \} \Big) \Big) \Big)
\end{multline}
\end{small}
where \( h_e^{(l)} \) represents the node embedding at layer \( l \), \( \mathcal{N} \) denotes the set of neighbors of \( e \), \( \text{AGG} \) is the \textit{mean} aggregation function, \( W^{(l)} \) is a learnable weight matrix, and \( \sigma \) is a non-linear activation function. This iterative process allows each node to incorporate information from an expanding receptive field, thus capturing both local and more global structural patterns within the graph. The final feature vector of the edge will be $\textbf{f}_e = h_e^L \in \mathbb{R}^{D}$.

This sampling and aggregation strategy allows the model to adapt to varying graph structures and sizes.  Furthermore, the inherent invariance properties of GNNs are crucial for modeling sewing patterns. Permutation invariance ensures consistent predictions regardless of edge indexing, while size invariance enables adaptation to patterns of varying complexity without manual padding as required in previous work~\cite{Chen_panelformer_2024, liu_towards_2023}.



\section{Differentiable Optimal Transport} \label{sec:optimal}
This section details the differentiable optimization strategy employed by \method for solving the assignment problem in garment edge matching. Leveraging Optimal Transport (OT) theory and the Sinkhorn algorithm, \method enables end-to-end differentiable learning, addressing the limitations of traditional non-differentiable methods, such as the Hungarian algorithm \cite{kuhn_hungarian_1955} and integer optimization techniques— which require separate optimization steps and struggle to adapt to dynamic data. See Supplementary Material for a complete explanation of the Sinkhorn algorithm 


\subsection{Partial Assignment Formulation}

Our goal is to estimate an assignment matrix $\mathbf{P} \in [0, 1]^{M \times M}$, that encodes whether two edges are stitched together or remain unpaired. 
Traditional assignment methods~\cite{kuhn_hungarian_1955, wolsey_integerprogramming_2000} typically require pairing every element; however, garment edge sewing requires a more flexible approach wherein certain edges may remain unstitched, necessitating a partial assignment strategy that allows for unmatched elements without compromising overall solution quality. 
This can be done by flexibilizing the constraints of $\mathbf{P}$ so that,
\begin{equation}
\mathbf{P} \mathbf{1}_{M} \leq \mathbf{1}_{M}, \quad
\mathbf{P}^{\top} \mathbf{1}_{M} \leq \mathbf{1}_{M},
\end{equation}
allowing us to pose the problem in a learning framework without requiring complex linear programming solvers~\cite{berthouzoz_parsing_2013}.
The assignment problem consists of maximizing a global score, 
$\sum_{i,j} \mathbf{C}_{i,j} \mathbf{P}_{i,j}$, where $\mathbf{C} \in \mathbb{R}^{M \times M}$ encodes the pairwise cost of an assignment between two stitching edges $i$ and $j$.

\paragraph{Score Prediction} 
We build the cost matrix $\mathbf{C}$ with all the possible pairings. 
As opposed to previous work~\cite{paul_superglue_2019}, which often frame the problem as bipartite matching of two independent disjoint sets, our formulation involves matching a single set with itself, where each sewing edge can be paired with any other except itself. This requires modifying the Sinkhorn algorithm to enforce structural constraints while maintaining probabilistic soft assignments, which we explain in Supplementary Material. 
To compute pairwise costs, we define the similarity between sewing edges using the inner product between the feature vectors of the edges:
\begin{equation}
\mathbf{C}_{i,j} = \langle \mathbf{f}_i, \mathbf{f}_j \rangle, \quad \forall (i, j) \in M \times M.
\end{equation}
where $\langle \cdot, \cdot \rangle$ denotes the inner product of vectors. Since the cost matrix is symmetric, the assignment process ensures global consistency across all subsets while preserving structural dependencies between sewing edges. 


\paragraph{Unstitched Edges}
To enable \method to suppress certain sewing edges when necessary, we introduce a dummy assignment mechanism that explicitly handles unmatched edges. This technique, common in graph matching \cite{paul_superglue_2019, deTone_superpoint_2017}, assigns unmatched elements to a \textit{dustbin} node, thereby avoiding forced and potentially incorrect assignments. 

To implement this mechanism, we augment the original cost matrix $\mathbf{C}$ by appending an additional row and column, resulting in an extended matrix $\bar{\mathbf{C}} \in \mathbb{R}^{(M+1) \times (M+1)}$. The extra row and column correspond to assignments to the bin, ensuring that unmatched edges are explicitly accounted for and are either assigned to a corresponding edge from another subset or placed in the dustbin. The point-to-bin and bin-to-bin scores are regulated by a single learnable parameter $z \in \mathbb{R}$, such that,
\begin{equation}
\bar{C}_{i, M+1} = \bar{C}_{M+1, j} = \bar{C}_{M+1, M+1} = z.
\end{equation}
This generalization to the matching problem results in a symmetric extended cost matrix.

\section{End-to-End Optimization}\label{sec:endtoend}

\subsection{Loss}\label{sec:loss}
The proposed architecture is fully differentiable, facilitating end-to-end training. We use ground truth matches and unmatched edges as supervision. We minimize the negative log-likelihood of the predicted assignment matrix, 
%
%
\begin{equation}
\text{Loss} = - \sum_{(i,j) \in \mathcal{A}} \log \bar{\mathbf{P}}_{i,j},
\end{equation}
where $\bar{\mathbf{P}}$ is the extended soft assignment matrix which includes unstitched and stitched edges, and $\mathcal{A}$
is the set of ground truth assignments, including both matched pairs $M$ and unstitched edges assigned to the dustbin.


\subsection{Final Hard Assignment}\label{sec:hardassignment} 

The previous optimization  process yields a soft-assignment matrix $\bar{\mathbf{P}}$ that contains two potential solutions. To derive a unique assignment while ensuring normalization, we compute the element-wise average of the upper and lower triangular components:
\begin{equation}\label{eq:average}
\mathbf{P}' = \frac{1}{2} (\bar{\mathbf{P}} + \bar{\mathbf{P}}^\top).  \end{equation}
 
This operation yields a probability distribution for each row (or column) of the symmetrized score matrix \(\mathbf{P}'\).  
To support multi-edge stitching, we proceed as follows: for each row we sort the probabilities in descending order and always keep the highest-scoring entry \(i^{*}=\arg\max_i p_{ij}\).

In addition, we retain every other entry whose probability satisfies \(p_{ij}\ge\tau_{\mathrm{multi}}\), where  \(\tau_{\mathrm{multi}}\)  is a predefined threshold that controls the number of allowable multi-edge connections.   
If no additional probabilities exceed this threshold, the assignment defaults to a single-edge match.
For further implementation details, see Section \ref{sec:exp_hyperparameters}.

\section{Multi-Edge Stitching Dataset} 

We build on the dataset \dataset~\cite{korosteleva_garmentcodedata_2024}, which contains over 128K sewing pattern samples spanning a wide range of designs, over 1.3 million panels (ranging from 2 to 37 panels per pattern, with an average of 10) and 4.8 million edges (averaging 10 edges per panel). Each garment pattern includes approximately 37 stitching connections, capturing the structural complexity typical of real-world garment construction.
This dataset, despite its scale, is restricted to one-to-one stitch connections —a simplification that fails to reflect the multi-edge stitching commonly required in practical manufacturing settings.
To address this limitation, we collaborated with a professional fashion designer, identifying two important oversimplifications of the patterns: 
\begin{itemize}[noitemsep,nosep]
	\item \textbf{Sleeve division:} Real-world sleeves are typically cut as a single panel. This is usually sewn to one cuff and attached simultaneously to both front and back torsos in a multi-to-one configuration. The original dataset, by contrast, represents sleeves as two mirrored halves with only one-to-one stitch annotations.
	\item \textbf{Upper-body division:} Industrial patterns often treat each torso (front or back) as a unified panel. However, the dataset splits both into separate pieces, introducing artificial seams that do not reflect the real ones.
\end{itemize}
%

%

\noindent Each pattern affected by the simplifications described previously was automatically detected and processed as follows. 
For patterns containing sleeves, we applied a three-step procedure: (1) Panel mirroring: the back-sleeve panel is reflected across its horizontal axis, and the back-cuff panel across its vertical axis; (2) Geometric merging: each mirrored sleeve and cuff half is merged with its front counterpart when their edge curvature and length are compatible; (3) Edge collapsing: adjacent edges in the merged panels are collapsed and concatenated, and the original stitch annotations are transferred to the resulting combined edges to encode multi-edge connections.
For the upper body, we merge front and back torso panels while preserving the original stitch maps.
This workflow yields over 18K updated patterns with more realistic multi-edge stitching relationships, released as the \newdataset dataset (where \textsc{M-E}\xspace stands for \textit{Multi-Edge}); see sample in Fig.~\ref{fig:new_pattern}. Full implementation details are provided in the Supplementary Material.
The dataset is randomly split into training, validation, and test sets in an 80-10-10 ratio.

%


\begin{figure}
	\centering
	\includegraphics[width=0.5\textwidth]{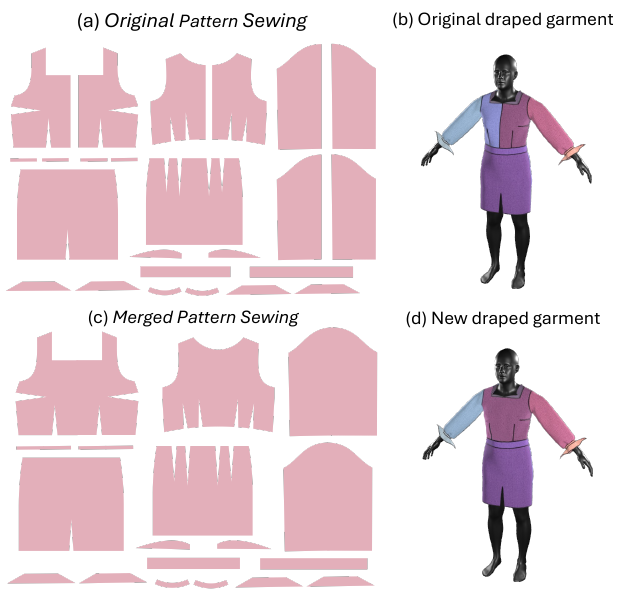}
	\caption{Comparison of sewing patterns and draped avatars before and after multi-edge merging. Top: original dataset with one-to-one stitches, showing artificial seam marks. Bottom: our Multi-Edge dataset with realistic multi-to-one stitches, producing more natural seams and garment reconstructions.
}

	\label{fig:new_pattern}
\end{figure}




\section{Implementation Details} \label{sec:exp_hyperparameters}

The \method pipeline uses a GraphSAGE-based GNN with $L = 5$ hidden layers, each with 512 neurons and ReLU activations, to process 24-dimensional edge features. Mean pooling aggregates neighbors, producing embeddings with $D = 128$ dimensions to capture local and global pattern relationships. The differentiable Sinkhorn algorithm performs $T$ = 100 iterations for partial assignment in stitching prediction. We set the multi‐edge threshold \(\tau_{\mathrm{multi}}=0.4\) based on ablation studies (Section~\ref{sec:ablation}), exploring the trade‐off among the metrics.
The whole pipeline is implemented in PyTorch and trained on an NVIDIA RTX 4090 GPU with a 0.001 learning rate over 18 epochs, achieving efficient convergence. A forward pass takes approximately 7 ms for a typical garment pattern.

\section{Evaluation}

In this section, we present quantitative metrics to evaluate the stitching prediction accuracy and robustness of our method, as well as ablation studies to analyze the impact of key components. It is important to note that, in neural state-of-the-art approaches such as SewFormer \cite{liu_towards_2023}, the embedding representation is inherently entangled with the learning operation. It is not possible to disentangle token learning from the stitching process without fundamentally modifying their original framework for comparison purposes. 
\subsection{Metrics}

\begin{figure*}[t]
    \centering    
    \includegraphics[width=0.8\textwidth]{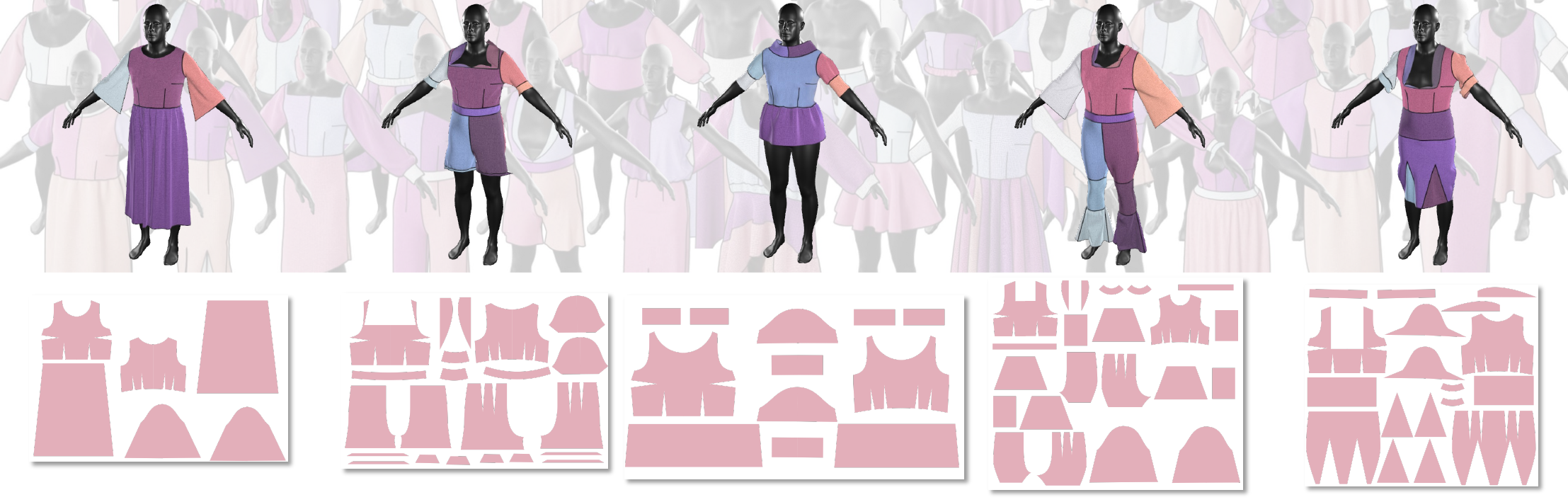}
    \caption{Random results showcasing actual patterns from the \newdataset.}
    \label{fig:dataset-results}
\end{figure*}

We evaluate \method's performance using seven metrics. 
%
Total Precision (TP) measures the proportion of correctly predicted stitch correspondences among all predicted edges, while Total Recall (TR) quantifies the fraction of ground-truth correspondences successfully identified. Total F1-score (TF1) balances these two measures, providing an overall metric of accuracy.
To specifically evaluate multi-edge stitching, we compute these metrics on the subset of multi-edge connections, yielding Multi-edge Precision (MEP), Multi-edge Recall (MER), and their harmonic mean, the Multi-edge F1-score (MEF1).
%

Finally, we introduce Garment Sewn Percentage (GSP) which is the ratio of correctly resolved patterns (\textit{i.e.}, those where all ground-truth stitch correspondences are predicted correctly) to the total number of test patterns.
GSP provides a practical measure of \method's effectiveness in real-world garment assembly. 
All metrics are computed from the  $M \times M$  hard assignment matrix produced by the optimizer, excluding edges assigned to the dustbin.

\subsection{Accuracy and Comparison}

\paragraph{Accuracy} We evaluate accuracy on the test set, obtaining 96.1\% TP, 96.3\% TR, and 96.2\% TF1; for multi‐edge stitching it achieves 80.4\% MEP and 89.5\% MER, yielding a 84.7\% MEF1. Overall, the model fully resolves 73.3\% of the test garment patterns (GSP). For comparison, on the original \dataset, which only contains one-to-one stitchings, \method reaches 97.19\% precision, 96.93\% recall, and 97.06\% F1-score, fully resolving 80.60\% of the test garment patterns (GSP). The higher performance on this dataset is expected, as it represents a simpler setting without multi-to-one cases, whereas \newdataset introduces more realistic stitching scenarios that are inherently more challenging.
Some fully assembled results are shown in Figure~\ref{fig:dataset-results}. We have observed that a significant proportion of failure cases happens when multiple panels share identical shapes, a common occurrence rendering them interchangeable. This performance degradation is attributed to mislabeling during training, where valid solutions are incorrectly classified as failures due to the use of correct stitches as ground truth.
\subsection{Ablation Studies}  \label{sec:ablation}

In this section, we analyze the impact of the most relevant decisions of the pipeline: the selection of geometric features, the size of the GNN, the number of iterations of the optimal transport problem, and the confidence threshold for the assignment.

\paragraph{Impact of the Geometric Features}  
We further analyze the contribution of the different geometric descriptors used in \method. 
In particular, we remove two components: (i) the \emph{Panel ID}, a unique identifier \(u\) assigned to each panel, and (ii) the topological properties. 
The results in Table~\ref{tab:ablation_features} show that eliminating Panel ID \(u\) degrades performance, especially in GSP, confirming that unique panel identifiers complement geometry by helping to disambiguate symmetric or repetitive edges. 
Removing topological properties, on the other hand, causes a sharp drop of nearly 18\% in TF1 and more than 40\% in GSP, underlining their importance for capturing global garment structure. 
Overall, these ablations demonstrate that the full feature set provides the most effective representation, and that both local shape descriptors and topological properties are crucial for accurate and robust stitching prediction.

\begin{table}[htb]
  \centering
  \small
  \renewcommand{\arraystretch}{1.1}
  \begin{tabular}{lcc}
    \toprule
    \textbf{Config.}         & \textbf{TF1 / MEF1} & \textbf{GSP} \\
    \midrule
    Full                     & \textbf{96.2 / 84.7} & \textbf{73.3} \\
    \textit{w/o} Panel ID    & 87.7 / 72.4          & 45.3 \\
    \textit{w/o} Topol. Prop.& 78.8 / 73.7          & 30.5 \\
    \bottomrule
  \end{tabular}
  \caption{Ablation study on the geometric descriptors. Both topological properties and the Panel ID (unique panel identifier) are crucial for stitching prediction performance.}
  \label{tab:ablation_features}
\end{table}

\paragraph{Impact of the Pipeline Design}  
We analyze the impact of the different design decisions of \method. We replace the GNN with a simple Multi-Layer Perceptron (MLP) using identical hyperparameters but without message aggregation. What we observe is a decrease of more than 2\% in TF1, 4\% in MEF1 and a drop of nearly 15\% in GSP. These results emphasize the importance of message passing for the complete garment assembly.
%


\begin{table}[htb]
  \centering
  \small
  \renewcommand{\arraystretch}{1.1}
  \begin{tabular}{lcccc}
    \toprule
    \textbf{Config.}   & \textbf{TP/MEP}  & \textbf{TR/MER}  & \textbf{TF1/MEF1} & \textbf{GSP} \\
    \midrule
    Full               & \textbf{96.1/80.4} & \textbf{96.3/89.5} & \textbf{96.2/84.7} & \textbf{73.3} \\
    \textit{w/o} GNN   & 93.2/71.2         & 94.2/92.9         & 93.7/80.6         & 58.6         \\
    \bottomrule
  \end{tabular}
  \caption{Ablation study results on the test set. Removing message aggregation (GraphSAGE) severely impacts performance, particularly GSP.}
  \label{tab:ablation}
\end{table}

\paragraph{Effect of GNN Design Choices}  
Table~\ref{tab:gnn-layers} shows the performance of several versions of the architecture of the GNN by varying the number of hidden layers and the aggregation operator.
The number of hidden layers determines the receptive field size so that a larger receptive field allows nodes to aggregate information from more distant neighbors potentially improving performance. However, excessive depth can lead to over-smoothing, where node embeddings become too similar, reducing model effectiveness.
This effect can be observed in all the metrics and mainly in the GSP, where the best result is obtained with 5 layers (our final model) and the worst results are obtained using 3 layers. 
Increasing the number of layers to seven results in a substantial drop in GSP (71.6\%), indicating that excessive information aggregation from neighbors, may dilute crucial local features. 
MEF1 follows a similar trend to TF1 and GSP, achieving its best results with a 5-layer GNN.

%

Additionally, we experimented with replacing \method's default \textsc{mean} aggregator with a \textsc{max} aggregator while keeping five hidden layers. As shown in Table \ref{tab:gnn-layers}, this alternative configuration yields similar results, achieving 96.1\% TF1 and 84.8\% MEF1. While \textsc{max} aggregation slightly reduces GSP compared to the mean-based approach, it maintains competitive performance, suggesting that the model is robust to different aggregation strategies.

\begin{table}[htb]
\centering
\renewcommand{\arraystretch}{1.1}
\begin{tabular}{lccc}
\toprule
\textbf{GNN layers} & \textbf{Total F1} & \textbf{Multi-edge F1} & \textbf{GSP} \\
\midrule
$L=3$ & 94.3 & 81.8 & 61.8 \\
\textbf{$L=5$} & \textbf{96.2} & \textbf{84.7} & \textbf{73.3} \\
$L=7$ & 96.0 & 78.9 &  71.6 \\
\hline
5 + Max  & 96.1 & 84.8 & 72.4 \\
\hline
\end{tabular}
\caption{Impact of GNN depth and aggregation method on stitching prediction performance. Increasing the number of layers improves performance up to $L=5$, but deeper networks ($L=7$) exhibit reduced GSP due to over-smoothing. Using max aggregation instead of mean results in similar performance, with a slight drop in recall.}
\label{tab:gnn-layers}
\end{table}

\paragraph{Impact of Sinkhorn Iterations ($T$)}
Table \ref{tab:sinkhorn-iterations} highlights the trade-off between accuracy and computational efficiency when varying the number of Sinkhorn iterations. 

\begin{table}[htb]
\centering
\renewcommand{\arraystretch}{1.1}
\begin{tabular}{lccc}
\toprule
\textbf{Configuration} & \textbf{TF1} & \textbf{MEF1} & \textbf{Time} \\
\midrule
$T=500$     & 97.1  & 84.5  &  26ms         \\
$T=100$     & 97.0  & 84.7  & \textbf{7ms}  \\
\hline
\end{tabular}
\caption{Effect of Sinkhorn iterations ($T$) on stitching prediction performance and inference time. 
}
\label{tab:sinkhorn-iterations}
\end{table}

Reducing Sinkhorn iterations from $T=500$ to $T=100$ cuts inference runtime by nearly $3\times$ with minimal impact on accuracy.

\paragraph{Ablation of the Multi-Edge Threshold \(\tau_{\mathrm{multi}}\)}

\begin{figure}
    \centering
    \includegraphics[width=0.7\linewidth]{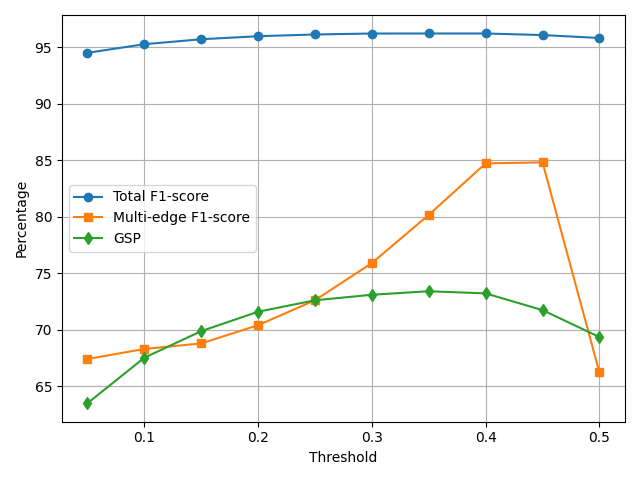}
    \caption{Impact of confidence threshold on \method's stitching prediction performance.}
    \label{fig:threshold-figure}
\end{figure}

To quantify how the multi-edge threshold \(\tau_{\mathrm{multi}}\) affects both total accuracy and multi-edge recovery, we swept \(\tau_{\mathrm{multi}}\) from 0.05 to 0.5 and measured the performance. Table~\ref{fig:threshold-figure} summarizes the results.
We observe that both TF1 and MEF1 reach their peak values when setting the threshold \(\tau_{\mathrm{multi}} = 0.4\). This confirms that a moderate threshold balances precision and recall, effectively capturing true multi-edge connections without introducing excessive noise. Notably, performance on multi-edge predictions drops sharply beyond \(\tau_{\mathrm{multi}} = 0.5\), as higher thresholds prevent assigning more than two edges to the same counterpart.  
Although we fix \(\tau_{\mathrm{multi}}\) at 0.4 in our experiments, this value can be adapted to new datasets depending on the expected frequency of multi-edge seams.


\new{
}

\section{Conclusions and Future Work}

We have introduced \method, the first fully automatic, end-to-end framework capable of inferring multi-edge stitching connections purely from geometric features. 
%
Thanks to a novel formulation leveraging GNN and differentiable optimal transport, our method achieves state-of-the-art accuracy, robustly handles pattern variability, and removes the need for semantic labels. Its streamlined design simplifies its application to diverse digital garment data models, making it well-suited for industrial manufacturing.
In addition, we also provide a novel dataset, \newdataset, an updated version of \dataset, enriched with multi-edge annotations, establishing a new benchmark for industrial applications.

While our approach achieves good and consistent results with the most common multiedge stitches (2-to-1), to improve generalization to real-world patterns, we plan to collect and expand the generation of samples on our real-world dataset including \(N\)-to-1 stitching connectivities. 
We expect that, by adjusting our multi-edge threshold $\tau_{\mathrm{multi}}$ the model will flexibly accommodate different types of multi-edge connections, ranging from simple two-to-one seams to more intricate configurations. Also, adaptive learning of this parameter could have a potentially high impact on improving robustness, as well as combining our automatic predictions with semantic labels, if available. 
Furthermore, additional topology preprocessing could be learned, to subdivide the input sewing edges and reduce the number ($N$) of simultaneous connections.

\method is, to our knowledge, the first automatic stitching method that works solely with simple 2D geometric shapes, requiring no custom semantics, annotations, or 3D data, while also explicitly supporting multi-edge stitching.

\section*{Acknowledgements}
Grant PID2021-122392OB-I00 funded by MCIN/AEI/ 10.13039/501100011033 and by ESF Investing in your future. It is also part of the project TaiLOR, CPP2021-008842 funded by MCIN/AEI/10.13039/501100011033 and the NextGenerationEU / PRTR programs. 

{
    \small
    \bibliographystyle{ieeenat_fullname}
    \bibliography{main}

@String(CVPR= {IEEE Conf. Comput. Vis. Pattern Recog.})

@String(ICCV= {Int. Conf. Comput. Vis.})

@String(ECCV= {Eur. Conf. Comput. Vis.})

@String(CVPR  = {CVPR})

@String(ICCV  = {ICCV})

@String(ECCV  = {ECCV})

@misc{nakayama_aipparel_2024,
	title = {{AIpparel}: {A} {Large} {Multimodal} {Generative} {Model} for {Digital} {Garments}},
	shorttitle = {{AIpparel}},
	url = {http://arxiv.org/abs/2412.03937},
	doi = {10.48550/arXiv.2412.03937},
	urldate = {2024-12-17},
	publisher = {arXiv},
	author = {Nakayama, Kiyohiro and Ackermann, Jan and Kesdogan, Timur Levent and Zheng, Yang and Korosteleva, Maria and Sorkine-Hornung, Olga and Guibas, Leonidas J. and Yang, Guandao and Wetzstein, Gordon},
	month = dec,
	year = {2024},
	note = {arXiv:2412.03937 [cs]},
}

@article{deTone_superpoint_2017,
  author       = {Daniel DeTone and
                  Tomasz Malisiewicz and
                  Andrew Rabinovich},
  title        = {SuperPoint: Self-Supervised Interest Point Detection and Description},
  journal      = {CoRR},
  volume       = {abs/1712.07629},
  year         = {2017},
  url          = {http://arxiv.org/abs/1712.07629},
  eprinttype    = {arXiv},
  eprint       = {1712.07629},
  bibsource    = {dblp computer science bibliography, https://dblp.org}
}

@article{paul_superglue_2019,
  author       = {Paul{-}Edouard Sarlin and
                  Daniel DeTone and
                  Tomasz Malisiewicz and
                  Andrew Rabinovich},
  title        = {SuperGlue: Learning Feature Matching with Graph Neural Networks},
  journal      = {CoRR},
  volume       = {abs/1911.11763},
  year         = {2019},
  url          = {http://arxiv.org/abs/1911.11763},
  eprinttype    = {arXiv},
  eprint       = {1911.11763},
  bibsource    = {dblp computer science bibliography, https://dblp.org}
}

@INPROCEEDINGS{eisenber_implicitsinkhorn_2022,
  author={Eisenberger, Marvin and Toker, Aysim and Leal-Taixé, Laura and Bernard, Florian and Cremers, Daniel},
  booktitle={2022 IEEE/CVF Conference on Computer Vision and Pattern Recognition (CVPR)}, 
  title={A Unified Framework for Implicit Sinkhorn Differentiation}, 
  year={2022},
  pages={499-508},
  doi={10.1109/CVPR52688.2022.00059}}

@article{berthouzoz_parsing_2013,
	title = {Parsing sewing patterns into {3D} garments},
	volume = {32},
	issn = {0730-0301, 1557-7368},
	url = {https://dl.acm.org/doi/10.1145/2461912.2461975},
	doi = {10.1145/2461912.2461975},
	language = {en},
	number = {4},
	urldate = {2024-05-20},
	journal = {ACM Transactions on Graphics},
	author = {Berthouzoz, Floraine and Garg, Akash and Kaufman, Danny M. and Grinspun, Eitan and Agrawala, Maneesh},
	month = jul,
	year = {2013},
	pages = {1--12},
}

@article{huamin_automatic_2024,
author = {Liu, Chen and Xu, Weiwei and Yang, Yin and Wang, Huamin},
title = {Automatic Digital Garment Initialization from Sewing Patterns},
year = {2024},
issue_date = {July 2024},
publisher = {Association for Computing Machinery},
address = {New York, NY, USA},
volume = {43},
number = {4},
issn = {0730-0301},
url = {https://doi.org/10.1145/3658128},
doi = {10.1145/3658128},
journal = {ACM Trans. Graph.},
month = jul,
articleno = {74},
numpages = {12}
}

@misc{peyré_computationaloptimaltransport_2020,
      title={Computational Optimal Transport}, 
      author={Gabriel Peyré and Marco Cuturi},
      year={2020},
      eprint={1803.00567},
      archivePrefix={arXiv},
      primaryClass={stat.ML},
      url={https://arxiv.org/abs/1803.00567}, 
}

@article{kuhn_hungarian_1955,
	title = {The {Hungarian} method for the assignment problem},
	volume = {2},
	copyright = {http://onlinelibrary.wiley.com/termsAndConditions\#vor},
	issn = {0028-1441, 1931-9193},
	url = {https://onlinelibrary.wiley.com/doi/10.1002/nav.3800020109},
	doi = {10.1002/nav.3800020109},
	language = {en},
	number = {1-2},
	urldate = {2025-02-18},
	journal = {Naval Research Logistics Quarterly},
	author = {Kuhn, H. W.},
	month = mar,
	year = {1955},
	pages = {83--97},
}

@article{sinkhorn_concerning_1967,
	title = {Concerning nonnegative matrices and doubly stochastic matrices},
	volume = {21},
	issn = {0030-8730, 0030-8730},
	url = {http://msp.org/pjm/1967/21-2/p14.xhtml},
	doi = {10.2140/pjm.1967.21.343},
	language = {en},
	number = {2},
	urldate = {2025-02-18},
	journal = {Pacific Journal of Mathematics},
	author = {Sinkhorn, Richard and Knopp, Paul},
	month = may,
	year = {1967},
	pages = {343--348},
}

@misc{dosovitskiy2021,
      title={An Image is Worth 16x16 Words: Transformers for Image Recognition at Scale}, 
      author={Alexey Dosovitskiy and Lucas Beyer and Alexander Kolesnikov and Dirk Weissenborn and Xiaohua Zhai and Thomas Unterthiner and Mostafa Dehghani and Matthias Minderer and Georg Heigold and Sylvain Gelly and Jakob Uszkoreit and Neil Houlsby},
      year={2021},
      eprint={2010.11929},
      archivePrefix={arXiv},
      primaryClass={cs.CV},
      url={https://arxiv.org/abs/2010.11929}, 
}

@InProceedings{Chen_panelformer_2024,
    author    = {Chen, Cheng-Hsiu and Su, Jheng-Wei and Hu, Min-Chun and Yao, Chih-Yuan and Chu, Hung-Kuo},
    title     = {Panelformer: Sewing Pattern Reconstruction From 2D Garment Images},
    booktitle = {Proceedings of the IEEE/CVF Winter Conference on Applications of Computer Vision (WACV)},
    month     = {January},
    year      = {2024},
    pages     = {454-463}
}

@inproceedings{Korosteleva_GarmentData_2021,
 author = {Korosteleva, Maria and Lee, Sung-Hee},
 booktitle = {Proceedings of the Neural Information Processing Systems Track on Datasets and Benchmarks},
 editor = {J. Vanschoren and S. Yeung},
 pages = {},
 title = {Generating Datasets of 3D Garments with Sewing Patterns},
 url = {https://datasets-benchmarks-proceedings.neurips.cc/paper/2021/file/013d407166ec4fa56eb1e1f8cbe183b9-Paper-round1.pdf},
 volume = {1},
 year = {2021}
}

@article{Korosteleva_NeuralTailor_2022,
   title={NeuralTailor: reconstructing sewing pattern structures from 3D point clouds of garments},
   volume={41},
   ISSN={1557-7368},
   url={http://dx.doi.org/10.1145/3528223.3530179},
   DOI={10.1145/3528223.3530179},
   number={4},
   journal={ACM Transactions on Graphics},
   publisher={Association for Computing Machinery (ACM)},
   author={Korosteleva, Maria and Lee, Sung-Hee},
   year={2022},
   month=jul, pages={1–16} }

@article{Korosteleva_GarmentCode_2023,
  author = {Korosteleva, Maria and Sorkine-Hornung, Olga},
  title = {{GarmentCode}: Programming Parametric Sewing Patterns},
  year = {2023},
  issue_date = {December 2023},
  publisher = {Association for Computing Machinery},
  address = {New York, NY, USA},
  volume = {42},
  number = {6},
  doi = {https://doi.org/10.1145/3618351},
  journal = {ACM Transaction on Graphics},
  note = {SIGGRAPH ASIA 2023 issue},
  numpages = {16}
}

@misc{korosteleva_garmentcodedata_2024,
	title = {{GarmentCodeData}: {A} {Dataset} of {3D} {Made}-to-{Measure} {Garments} {With} {Sewing} {Patterns}},
	shorttitle = {{GarmentCodeData}},
	url = {http://arxiv.org/abs/2405.17609},
	urldate = {2024-06-06},
	publisher = {arXiv},
	author = {Korosteleva, Maria and Kesdogan, Timur Levent and Kemper, Fabian and Wenninger, Stephan and Koller, Jasmin and Zhang, Yuhan and Botsch, Mario and Sorkine-Hornung, Olga},
	month = may,
	year = {2024},
	note = {arXiv:2405.17609 [cs]},
}

@article{Loper2015,
author = {Loper, Matthew and Mahmood, Naureen and Romero, Javier and Pons-Moll, Gerard and Black, Michael J.},
title = {SMPL: a skinned multi-person linear model},
year = {2015},
issue_date = {November 2015},
publisher = {Association for Computing Machinery},
address = {New York, NY, USA},
volume = {34},
number = {6},
issn = {0730-0301},
url = {https://doi.org/10.1145/2816795.2818013},
doi = {10.1145/2816795.2818013},
journal = {ACM Trans. Graph.},
month = oct,
articleno = {248},
numpages = {16}
}

@inproceedings{Umetani2011,
author = {Umetani, Nobuyuki and Kaufman, Danny M. and Igarashi, Takeo and Grinspun, Eitan},
title = {Sensitive couture for interactive garment modeling and editing},
year = {2011},
isbn = {9781450309431},
publisher = {Association for Computing Machinery},
address = {New York, NY, USA},
url = {https://doi.org/10.1145/1964921.1964985},
doi = {10.1145/1964921.1964985},
booktitle = {ACM SIGGRAPH 2011 Papers},
articleno = {90},
numpages = {12},
location = {Vancouver, British Columbia, Canada},
series = {SIGGRAPH '11}
}

@article{liu_towards_2023,
	title = {Towards {Garment} {Sewing} {Pattern} {Reconstruction} from a {Single} {Image}},
	volume = {42},
	issn = {0730-0301, 1557-7368},
	url = {http://arxiv.org/abs/2311.04218},
	doi = {10.1145/3618319},
	number = {6},
	urldate = {2023-12-14},
	journal = {ACM Transactions on Graphics},
	author = {Liu, Lijuan and Xu, Xiangyu and Lin, Zhijie and Liang, Jiabin and Yan, Shuicheng},
	month = dec,
	year = {2023},
	note = {arXiv:2311.04218 [cs]},
	pages = {1--15},
}

@misc{he_dresscode_2024,
	title = {{DressCode}: {Autoregressively} {Sewing} and {Generating} {Garments} from {Text} {Guidance}},
	shorttitle = {{DressCode}},
	url = {http://arxiv.org/abs/2401.16465},
	doi = {10.1145/3658147},
	urldate = {2024-05-27},
	author = {He, Kai and Yao, Kaixin and Zhang, Qixuan and Liu, Lingjie and Yu, Jingyi and Xu, Lan},
	month = may,
	year = {2024},
	note = {arXiv:2401.16465 [cs]},
}

@misc{hamilton_inductive_2018,
      title={Inductive Representation Learning on Large Graphs}, 
      author={William L. Hamilton and Rex Ying and Jure Leskovec},
      year={2018},
      eprint={1706.02216},
      archivePrefix={arXiv},
      primaryClass={cs.SI},
      url={https://arxiv.org/abs/1706.02216}, 
}

@misc{scarselli_gnn_2009,
  author={Scarselli, Franco and Gori, Marco and Tsoi, Ah Chung and Hagenbuchner, Markus and Monfardini, Gabriele},
  journal={IEEE Transactions on Neural Networks}, 
  title={The Graph Neural Network Model}, 
  year={2009},
  volume={20},
  number={1},
  pages={61-80},
  doi={10.1109/TNN.2008.2005605}
}

@inProceedings{Zanfir_2018,
author = {Zanfir, Andrei and Sminchisescu, Cristian},
title = {Deep Learning of Graph Matching},
booktitle = {Proceedings of the IEEE Conference on Computer Vision and Pattern Recognition (CVPR)},
month = {June},
year = {2018}
}

@inProceedings{li19,
  title = 	 {Graph Matching Networks for Learning the Similarity of Graph Structured Objects},
  author =       {Li, Yujia and Gu, Chenjie and Dullien, Thomas and Vinyals, Oriol and Kohli, Pushmeet},
  booktitle = 	 {Proceedings of the 36th International Conference on Machine Learning},
  pages = 	 {3835--3845},
  year = 	 {2019},
  editor = 	 {Chaudhuri, Kamalika and Salakhutdinov, Ruslan},
  volume = 	 {97},
  series = 	 {Proceedings of Machine Learning Research},
  month = 	 {09--15 Jun},
  publisher =    {PMLR},
  url = 	 {https://proceedings.mlr.press/v97/li19d.html},
}

@InProceedings{Wang_2019,
author = {Wang, Runzhong and Yan, Junchi and Yang, Xiaokang},
title = {Learning Combinatorial Embedding Networks for Deep Graph Matching},
booktitle = {Proceedings of the IEEE/CVF International Conference on Computer Vision (ICCV)},
month = {October},
year = {2019}
}

@InProceedings{Sarlin2020,
author = {Sarlin, Paul-Edouard and DeTone, Daniel and Malisiewicz, Tomasz and Rabinovich, Andrew},
title = {SuperGlue: Learning Feature Matching With Graph Neural Networks},
booktitle = {Proceedings of the IEEE/CVF Conference on Computer Vision and Pattern Recognition (CVPR)},
month = {June},
year = {2020}
}

@inbook{Rolinek2020,
   title={Deep Graph Matching via Blackbox Differentiation of Combinatorial Solvers},
   ISBN={9783030586041},
   ISSN={1611-3349},
   url={http://dx.doi.org/10.1007/978-3-030-58604-1_25},
   DOI={10.1007/978-3-030-58604-1_25},
   booktitle={Computer Vision – ECCV 2020},
   publisher={Springer International Publishing},
   author={Rolínek, Michal and Swoboda, Paul and Zietlow, Dominik and Paulus, Anselm and Musil, Vít and Martius, Georg},
   year={2020},
   pages={407–424} }

@article{Wang2021QAP,
  author={Wang, Runzhong and Yan, Junchi and Yang, Xiaokang},
  journal={IEEE Transactions on Pattern Analysis and Machine Intelligence}, 
  title={Neural Graph Matching Network: Learning Lawler’s Quadratic Assignment Problem With Extension to Hypergraph and Multiple-Graph Matching}, 
  year={2022},
  volume={44},
  number={9},
  pages={5261-5279},
  doi={10.1109/TPAMI.2021.3078053}}

@InProceedings{Wang2023,
    author    = {Wang, Runzhong and Guo, Ziao and Jiang, Shaofei and Yang, Xiaokang and Yan, Junchi},
    title     = {Deep Learning of Partial Graph Matching via Differentiable Top-K},
    booktitle = {Proceedings of the IEEE/CVF Conference on Computer Vision and Pattern Recognition (CVPR)},
    month     = {June},
    year      = {2023},
    pages     = {6272-6281}
}

@inproceedings{ratnayaka2025,
title={Learning Partial Graph Matching via Optimal Partial Transport},
author={Gathika Ratnayaka and James Nichols and Qing Wang},
booktitle={The Thirteenth International Conference on Learning Representations},
year={2025},
url={https://openreview.net/forum?id=uDXFOurrHM}
}

@incollection{wolsey_integerprogramming_2000,
title = {Integer programming},
pages = {273-285},
booktitle = {IEE Transactions 32},
author = {Wolsey, L},
year = {2000},
}

@misc{zhou2020continuityrotationrepresentationsneural,
      title={On the Continuity of Rotation Representations in Neural Networks}, 
      author={Yi Zhou and Connelly Barnes and Jingwan Lu and Jimei Yang and Hao Li},
      year={2020},
      eprint={1812.07035},
      archivePrefix={arXiv},
      primaryClass={cs.LG},
      url={https://arxiv.org/abs/1812.07035}, 
}

@misc{wang20244ddress4ddatasetrealworld,
      title={4D-DRESS: A 4D Dataset of Real-world Human Clothing with Semantic Annotations}, 
      author={Wenbo Wang and Hsuan-I Ho and Chen Guo and Boxiang Rong and Artur Grigorev and Jie Song and Juan Jose Zarate and Otmar Hilliges},
      year={2024},
      eprint={2404.18630},
      archivePrefix={arXiv},
      primaryClass={cs.CV},
      url={https://arxiv.org/abs/2404.18630}, 
}

@misc{li2025garmagenetmultimodalgenerativeframework,
      title={GarmageNet: A Multimodal Generative Framework for Sewing Pattern Design and Generic Garment Modeling}, 
      author={Siran Li and Chen Liu and Ruiyang Liu and Zhendong Wang and Gaofeng He and Yong-Lu Li and Xiaogang Jin and Huamin Wang},
      year={2025},
      eprint={2504.01483},
      archivePrefix={arXiv},
      primaryClass={cs.GR},
      url={https://arxiv.org/abs/2504.01483}, 
}

@misc{lu2023jigsawlearningassemblemultiple,
      title={Jigsaw: Learning to Assemble Multiple Fractured Objects}, 
      author={Jiaxin Lu and Yifan Sun and Qixing Huang},
      year={2023},
      eprint={2305.17975},
      archivePrefix={arXiv},
      primaryClass={cs.CV},
      url={https://arxiv.org/abs/2305.17975}, 
}
}

\end{document}


\title{\method: A Geometric Approach to Stitching Prediction with Graph Neural Networks 

Supplementary Material}

\maketitle
\thispagestyle{empty}
\appendix


In this Supplementary Material, we provide additional details to support the main contributions of the paper. These materials are intended to enhance reproducibility and offer a deeper understanding of key design choices:

\begin{tabbing}
\hspace{1.5em}\=\hspace{0.8\linewidth}\kill
A.\> Details of Geometric Edge Features        \\
B.\> Details of Curated Dataset Extension      \\
C.\> Sinkhorn Algorithm                        \\
\end{tabbing}

\section{Details of Geometric Edge Features}

Each garment is represented by a set of disconnected 2D panels. A panel is a closed polygon defined by \(N\) edges, \(E=\{\edge_{j}\}_{j=1}^N\), where each edge may be a straight segment, circular arc, quadratic Bézier spline, or a cubic Bézier spline.
Additionally, we introduce B-splines to represent edges created during panel merge (e.g., in sleeve reconstruction). These are formed by concatenating two Bézier splines using their control points and midpoint, allowing the modeling of more complex edge shapes.

To capture the geometric and structural variability of these edges, we extract 22 raw features per edge. These include both local shape descriptors and topological properties. A complete list of these features, along with their definitions, is provided in Table~\ref{tab:edge-features}. Finally, these features are preprocessed into a 24-dimensional edge vector.

\begin{table*}[htb]
    \centering
    \small
    \renewcommand{\arraystretch}{1.1}
    \begin{tabular}{lll}
        \toprule
        \textbf{Geometric Features} & \textbf{Feature} & \textbf{Description} \\
        \midrule
        \multirow{9}{*}{\textbf{Local Shape}} 
            & Start vertex \((x_0,y_0)\)    & Coordinates of the edge’s starting point \\
            & End vertex \((x_1,y_1)\)      & Coordinates of the edge’s ending point  \\
            & Length \(l\)                  & Euclidean distance between start and end \\
            & Orientation \((o_x,o_y)\)     & Unit vector between the initial and end edge point  \\
            & Curvature type \(k_t\)        & \begin{tabular}[t]{@{}l@{}}
            \(\{0\dots5\}\): straight segment, circular arcs quadratic Bézier,\\  
            cubic Bézier and B-Splines \\  
          \end{tabular} \\
            & Curvature params \(k=(k_1,\dots,k_{10})\) &
        \begin{tabular}[t]{@{}l@{}}
            All zeros for straight segment \\ 
            \((r,d,\theta,0,\dots,0)\) for circular arc \\
            \((q^1_x,q^1_y,0,\dots,0)\) for quadratic Bézier \\  
            \((q^1_x,q^1_y,q^2_x,q^2_y,0,\dots,0)\) for cubic Bézier \\
            \((q^1_x,q^1_y,q^2_x,q^2_y,c_x, c_y, q^3_x,q^3_y,q^4_x,q^4_y)\) for B-splines
            
          \end{tabular} \\
        \midrule
        \multirow{4}{*}{\textbf{Topological Properties}}
            & Interior angle left  $\alpha_l$ & Interior angle with left neighbor edge (radians) \\
            & Interior angle right $\alpha_r$ & Interior angle with right neighbor edge (radians) \\
            & Edge count \(N\)                  & Total number of edges in panel \\
            & Panel ID \(u\)                    & Unique identifier for this panel \\
        \bottomrule
    \end{tabular}
    \caption{Geometric edge features composed of Local Shape descriptor and Topological properties.}
    \label{tab:edge-features}
\end{table*}

\section{Detailed \newdataset}\label{sec:dataset_supp}

We start by selecting every pattern in \dataset that contains at least one sleeve. The extension pipeline then proceeds in three short stages:

\begin{enumerate}
  \item \textbf{Panel mirroring:} The back-sleeve panel is reflected on its horizontal axis.

  \item \textbf{Geometric merging:} The mirrored back sleeve is compared with the front sleeve; if the length of the edge and curvature match, both halves are merged into a single sleeve panel.

  \item \textbf{Edge collapsing:} Adjacent edges are merged, and their stitch labels are mapped to multi-edge annotations. Straight and coplanar edges are collapsed into one longer segment. Quadratic or cubic Bézier edges are concatenated by joining control points to form a B-spline, preserving curvature continuity. After merging, all original one-to-one annotations along the collapsed or concatenated edges are reassigned to the new unified edge. This ensures that every multi-edge relationship is mapped.

\end{enumerate}
We apply the mirroring on its vertical axis, fusion, and stitch-update steps to each sleeve cuff and then merge the front and back upper body panels into unified pieces, keeping all original stitch annotations.

This pipeline produces 18 003 patterns with merged sleeves and sleeve cuffs, and unified torsos. This increases realism and structural complexity. The resulting dataset provides a stronger benchmark for training and evaluating multi-edge stitch-prediction models.

\section{Sinkhorn Algorithm} 

\label{sec:sinkhorn}

We adopt the entropic optimal transport (OT) framework \cite{peyré_computationaloptimaltransport_2020}, solved efficiently via the Sinkhorn algorithm \cite{sinkhorn_concerning_1967, eisenber_implicitsinkhorn_2022}. This approach introduces an entropy term to the standard OT cost, resulting in a smooth and differentiable objective. This regularization is crucial for stable and efficient computation of transport plans, which we utilize for \method.
Formally, the objective is:
\begin{equation}
OT_\varepsilon (a,b) = \min_{ \mathbf{P} \in U(a,b)} \langle C,  \mathbf{P} \rangle - \varepsilon H( \mathbf{P}).
\end{equation}
where $H( \mathbf{P})$ represents the entropy of the assignment matrix $ \mathbf{P}$, and $\epsilon > 0$ regulates the strength of the entropy-based smoothing, and $U(a,b)$ is the set of feasible transport plans with given marginals $a$ and $b$. In our case, \(a = b \in \mathbb{R}^M\) are uniform distributions over the \(M\) panel edges.

The Sinkhorn algorithm efficiently computes a doubly stochastic approximation to the optimal transport plan via iterative row and column normalization. The entropic regularization introduced by this method provides crucial differentiability, ensuring that small perturbations in the cost matrix or marginals result in predictable gradient updates. This property is essential for end-to-end training of our network. By utilizing a Sinkhorn-based layer, we leverage gradient-based optimization for partial matching, effectively bypassing the challenges posed by non-differentiable combinatorial solvers.


{
    \small
    \bibliographystyle{ieeenat_fullname}
    \bibliography{main}
}